%% file: main.tex
\documentclass{article}

\usepackage[preprint]{neurips_2025}

\usepackage[utf8]{inputenc} 
\usepackage[T1]{fontenc}    
\usepackage{hyperref}       
\usepackage{url}            
\usepackage{booktabs}       
\usepackage{amsfonts}       
\usepackage{nicefrac}       
\usepackage{microtype}      
\usepackage{xcolor}         

\usepackage[compatibility=false]{caption}
\usepackage{multirow}
\usepackage{graphicx}
\usepackage{amsmath}
\usepackage{enumitem}
\usepackage{wrapfig}
\usepackage{subcaption}
\usepackage{float}
\usepackage{pifont}

\title{Hierarchical Self-Prompting SAM: A Prompt-Free Medical Image Segmentation Framework}

\author{
    \textbf{Mengmeng Zhang}$^{1,2}$ \ \ \ \ 
    \textbf{Xingyuan Dai}$^{1}$ \ \ \ \ 
    \textbf{Yicheng Sun}$^{1,2}$\ \ \ \
    \textbf{Jing Wang}$^{1}$ \\
    \textbf{Yueyang Yao}$^{1,2}$\ \ \ \
    \textbf{Xiaoyan Gong}$^{1}$ \ \ \ \ 
    \textbf{Fuze Cong}$^{3}$ \ \ \ \ 
    \textbf{Feiyue Wang}$^{1}$ \ \ \ \ 
    \textbf{Yisheng Lv}$^{1,2}$\thanks{Corresponding author.}  \\ 
    \\
    $^{1}$State Key Laboratory of Multimodal Artificial Intelligence System,\\
    Institute of Automation, Chinese Academy of Science, China \\
    $^{2}$School of Artificial Intelligence, University of Chinese Academy of Science, China \\
    $^{3}$Department of Radiology, Peking Union Medical College Hospital, \\Peking Union Medical College, Chinese Academy of Medical Sciences, Beijing 100730, China\\
    \texttt{\{zhangmengmeng2022, xingyuan.dai, sunyicheng2025\}@ia.ac.cn} \\
    \texttt{\{wangjing2014, xiaoyan.gong, feiyue.wang, yisheng.lv\}@ia.ac.cn}\\
    \texttt{yueyaoyang23@mails.ucas.ac.cn}\\
    \texttt{fuzecong@hotmail.com}
}

\begin{document}

\maketitle
\input{secs/abstract}
\input{secs/introduction}
\input{secs/related_work}
\input{secs/method}
\input{secs/experiment}
\input{secs/conclusion}


\medskip

{
\small
\bibliographystyle{plain}
\bibliography{main}
}






\end{document}

%% file: secs/abstract.tex
\begin{abstract}
  Although the Segment Anything Model (SAM) is highly effective in natural image segmentation,  it requires dependencies on prompts, which limits its applicability to medical imaging where manual prompts are often unavailable. Existing efforts to fine-tune SAM for medical segmentation typically struggle to remove this dependency. We propose Hierarchical Self-Prompting SAM (HSP-SAM), a novel self-prompting framework that enables SAM to achieve strong performance in prompt-free medical image segmentation. Unlike previous self-prompting methods that remain limited to positional prompts similar to vanilla SAM, we are the first to introduce learning abstract prompts during the self-prompting process. This simple and intuitive self-prompting framework achieves superior performance on classic segmentation tasks such as polyp and skin lesion segmentation, while maintaining robustness across diverse medical imaging modalities. Furthermore, it exhibits strong generalization to unseen datasets, achieving improvements of up to 14.04\% over previous state-of-the-art methods on some challenging benchmarks. These results suggest that abstract prompts encapsulate richer and higher-dimensional semantic information compared to positional prompts, thereby enhancing the model’s robustness and generalization performance. All models and codes will be released upon acceptance.
  
\end{abstract}


%% file: secs/introduction.tex
\section{Introduction}

Medical image segmentation is critical for clinical diagnosis and treatment planning by enabling precise delineation of anatomical structures and lesions. Due to the traditionally time-consuming and labor-intensive manual annotation by radiologists, there has been an increasing demand for efficient and precise automated annotation solutions. Recently, researchers have made significant progress in the field of medical image segmentation \cite{wang2022medical, chen2024towards, jiang2024zept}. Traditional task-specific small models often require more specialized and intricate designs to adapt to different tasks, significantly increasing the difficulty of deployment and integration with other tasks. The emergence of general-purpose segmentation models, with their powerful generalization capabilities combined with the pre-training and fine-tuning paradigm, allows these large models to be easily adapted to various downstream tasks.

Segmentation Anything Module (SAM) \cite{sam} as a general-purpose segmentation model for natural images, demonstrates exceptional interactive segmentation capabilities due to its training on the SA-1B dataset. However, SAM's reliance on prompts poses challenges in medical segmentation. Since vanilla SAM relies on positional prompts, whereas medical imaging fundamentally focuses on lesion detection. An ideal medical segmentation system must automate both lesion localization and segmentation. Some work has attempted to apply SAM to medical image segmentation through fine-tuning, but many studies still rely on manual prompts\cite{ma2024segment,Med-SA}. In addition, some methods attempt to compensate for the loss of performance without manual prompts\cite{huang2024segment, ma2024segment} by introducing data-specific structures\cite{self-prompt_sam, masksam, MA-SAM}, which in turn limit the segmentation framework's applicability across diverse imaging modalities. 
\begin{figure}[t]
    \centering
    \includegraphics[width=1\linewidth]{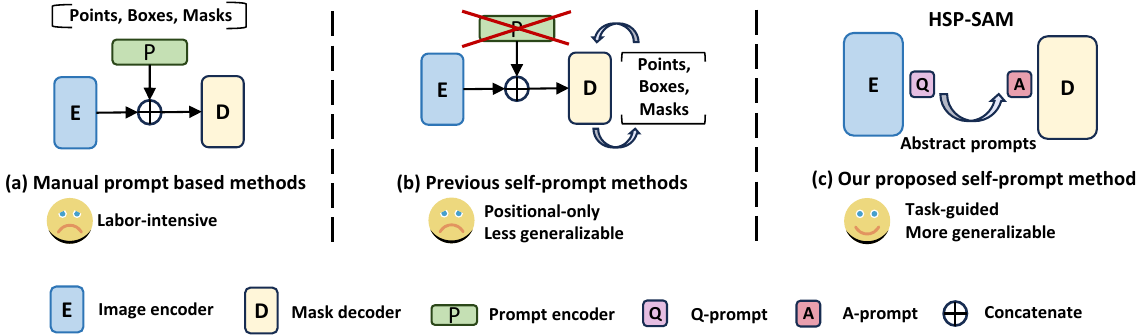}
    \caption{\textbf{Previous Methods \emph{vs.} HSP-SAM.} (a)illustrates interactive segmentation methods that rely on manual prompt inputs. (b)shows prior self-prompting approaches still constrained by positional prompts, lacking generalization. (c)presents our proposed self-prompting method that generates abstract task-guided prompts, achieving strong generalizability across modalities.}
    \label{fig:comperison}
\end{figure}
We argue that such adaptations reveal a fundamental distrust in the original SAM image encoder’s ability to generalize across medical imaging tasks. 
In practice, to achieve network-level inference speed, SAM trains a highly capable image encoder, and constructs extremely lightweight prompt and mask decoders. 
This design ensures that SAM's image encoder possesses strong feature extraction and generalization capabilities but makes SAM heavily rely on manual prompt inputs. 
Therefore, the main obstacle in transferring SAM to medical imaging tasks stems from its strong dependence on manual prompts rather than limitations in its image encoder's feature extraction capability.

We observe that SAM relies on manual prompt inputs to specify segmentation targets, as its original design uses point, box, and mask prompts to provide explicit localization information. However, we raise doubts about this prompting method. \textbf{Is positional information the only means to guide SAM's segmentation?} We believe this is not the case. At its core, SAM requires task guidance, and positional prompts are only one possible form, which may not be the most effective. While positional prompts offer the most direct form of task guidance, it does not align with the requirements of batch automated segmentation. In automated scenarios, guiding the model through generalized task descriptions—rather than providing target-specific positional prompts—is a more suitable and scalable solution. 

Considering the current challenges of adapting SAM to the medical domain, we propose Hierarchical Self-prompting SAM(HSP-SAM) to generate abstract task-guided prompts by self-prompting. This enables a prompt-free, generalizable segmentation framework for medical imaging. Fig. 1 illustrates the difference between methods relying on positional prompts and our proposed abstract self-prompting approach.
Specifically, we design the corresponding prompts for the image encoder and mask decoder, forming Q\&A prompt pairs. These paired prompts are derived through mapping transformations, guiding the encoder to discover key features and assisting the decoder in interpreting critical information. Through this mechanism, the Q\&A prompt pairs move beyond focusing on individual segmentation targets within a single input and instead capture a broader understanding of the overall segmentation task, providing more generalized task guidance for the model. Additionally, we designed a U-shaped structure to fuse high and low-dimensional features, enhancing the model's segmentation performance. The overall architecture of HSP-SAM is depicted in Fig. 2.

We conducted extensive experiments across diverse medical imaging modalities. The results demonstrate that HSP-SAM achieves superior robustness and generalization, consistently outperforming other SAM-based medical segmentation models and even surpassing some specialized models on challenging datasets.
In summary, our contributions are as follows.
\begin{itemize}[itemsep=0pt, parsep=0pt]
    \item Our findings suggest that SAM depends primarily on "task-guided prompts" rather than solely on "positional prompts."
    \item We find that abstract task-guided prompts, compared to positional prompts, enable the model to achieve a higher-level understanding of segmentation tasks, leading to improved generalization performance.
    \item We propose a self-prompting approach based on Q\&A prompt pairs to generate abstract task-guidance prompts, enabling the prompt-free transfer of SAM to medical image segmentation.
\end{itemize}

%% file: secs/related_work.tex
\section{Related Work}
Our method builds upon prompt learning and incorporates elements from traditional segmentation models. We also conduct extensive comparisons with SAM-based medical segmentation methods. Therefore, this section briefly reviews prompt learning, traditional segmentation approaches, and SAM-based segmentation methods.

\subsection{Prompt Learning}
Prompt Learning was initially proposed in the NLP field \cite{ding2021openprompt, gao2020making,jiang2020can} and later adopted in vision-language (V-L) and vision-only models \cite{zhou2022conditional,yan2023prompt, wang2024mcpl, wang2024vilt}. Originally, Prompt Learning refers to the method of manually designing or automatically learning prompts during fine-tuning to help text prompts, serving as instructions, better assist the text encoder in understanding tasks. This approach primarily involves replacing part of the original input text prompt \cite{yang2024text, coop} or adding additional prompts to the existing text prompt\cite{Wang_2022_CVPR,lei2024prompt}. As a Parameter Efficient Fine-Tuning (PEFT) method, prompt Learning is characterized by its simplicity and efficiency, leading to its gradual adoption in other domains. Moreover, the role of prompt learning continues to expand. For instance, MaPLe\cite{khattak2023maple} facilitates the alignment between the text and vision branches in V-L models through prompt learning, enabling CLIP to efficiently align text and images in downstream few-shot tasks. Inspired by MaPLe, our work also employs prompt learning to establish Q\&A prompt pairs between the encoder and decoder, thereby helping HSP-SAM adaptively discover abstract task-guided prompts.

\subsection{Traditional Segmentation Methods}
The emergence of U-Net\cite{u-net} marked a milestone in medical image segmentation, as it connected encoder and decoder features via skip connections, achieving the fusion of high- and low-dimensional features from input images and significantly improving segmentation performance in medical imaging. Subsequent works\cite{oktay2018Attention,isensee2021NnUNet, zhou2018UNet, milletari2016VNet} based on U-Net demonstrated the effectiveness of the U-shaped structure for medical images with blurred boundaries, low contrast, and irregularly shaped segmentation targets. With the rise of transformer architectures in computer vision, many studies have attempted to integrate the U-shaped structure into transformers\cite{vm-unet,swin-unetr}. These traditional segmentation models have validated the efficacy of the U-shaped structure for medical image segmentation, which is why our method also adopts this architecture in its design.

\subsection{SAM based Segmentation Methods}
Following the release of SAM, significant efforts have been made to adapt it for medical image segmentation by fine-tuning. For example, MedSAM\cite{ma2024segment} retrains SAM on an extensive medical dataset to mitigate the domain gap between natural and medical images; SAM-Med2D\cite{sammed2d} integrates adapters with 16 million multimodal medical images to achieve unified medical image segmentation; MA-SAM\cite{MA-SAM} targets 3D medical data, leveraging contextual information from adjacent slices to enhance SAM’s capabilities in volumetric imaging.

However, these approaches either still rely on external prompts for localization or are sensitive to the data structure. As a result, there has been growing interest in adapting SAM to medical image segmentation in a prompt-free and data-agnostic manner. For instance, SAM-SP\cite{zhou2024SAMSP} introduces a self-distillation strategy, generating bounding box prompts from self-produced masks to iteratively refine segmentation outputs, thereby enabling self-prompting. H-SAM\cite{h-sam} mitigates pixel imbalance by injecting Gaussian noise into inputs, improving prompt-free transfer performance on downstream tasks. ESP-MedSAM\cite{esp-sam} further boosts prompt-free transfer by introducing knowledge distillation and modality decoupling during decoding.

Unlike previous self-prompting methods still constrained by positional prompts, we propose HSP-SAM, which constructs abstract task-guided prompts through hierarchical Q\&A prompt pairs. These prompts transcend localization cues, allowing HSP-SAM to achieve superior generalization across multimodal medical datasets.

%% file: secs/method.tex
\section{Hierarchical Self-Prompting SAM}
\begin{figure}[!t]
    \centering
    \includegraphics[width=1.0\linewidth]{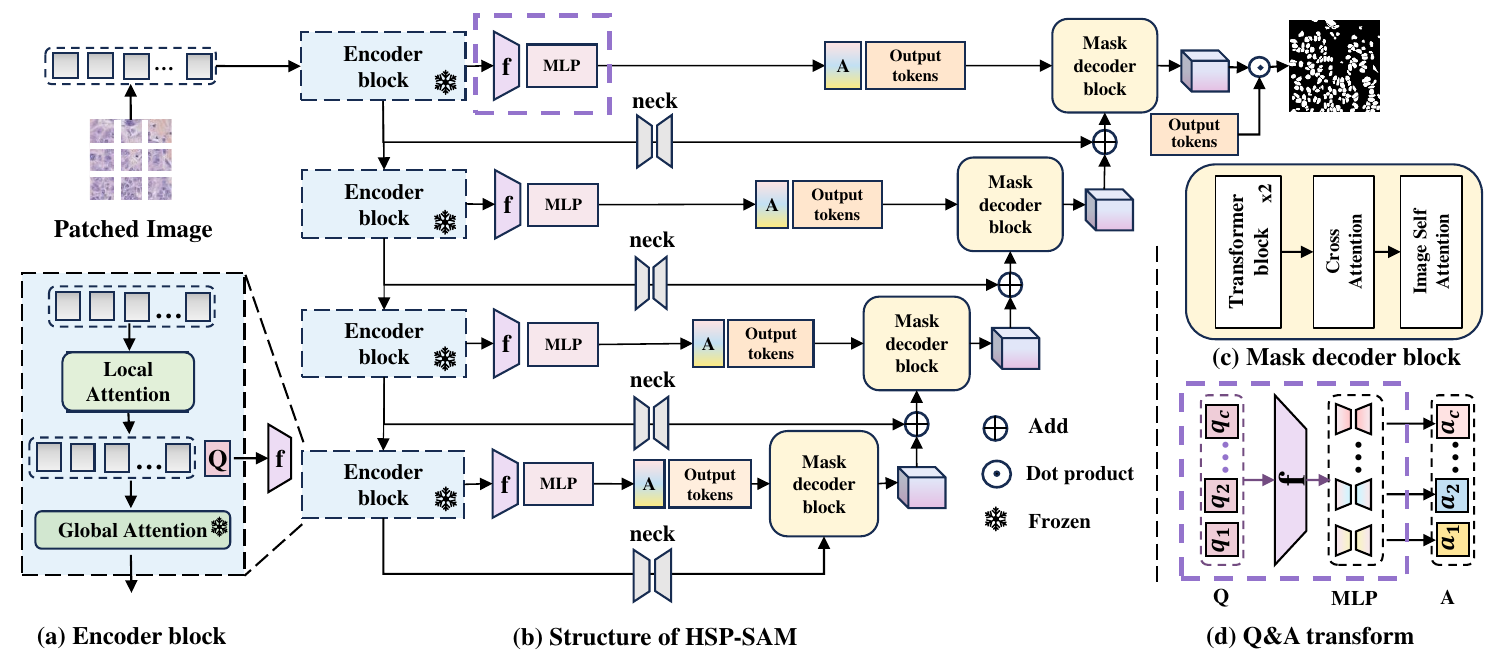}
    \caption{\textbf{Overview of HSP-SAM.} Notably, during training, all parameters of the image encoder are frozen, while the remaining components are trainable. \textbf{(d)} illustrates the transformation from $Q$ prompt to $A$ prompt, where all $q$ prompts are first processed through a shared mapping function $f$, and then further transformed into $a$ prompts by task-specific MLP modules.}
    \label{fig:enter-label}
\end{figure}

\subsection{Overview of HSP-SAM}
As illustrated in Figure 2, our proposed HSP-SAM aims to enhance the segmentation performance of the SAM on medical images without the need for manual prompts. HSP-SAM is a prompt-free medical image segmentation framework that is agnostic to input data modalities, enabling straightforward adaptation of SAM to various medical datasets. To achieve prompt-free, we introduce three key modifications in HSP-SAM: LoRA-based fine-tuning, a Q\&A-based self-prompt module, and a Hierarchical feature fusion module.

\subsection{LoRA-based Fine-tuning }
SAM is pre-trained on natural images that are substantially different from medical images. Despite this, SAM’s image encoder captures rich fundamental features, making it important to retain its learned knowledge. Fully fine-tuning the image encoder is computationally expensive, as SAM's approximately 1 billion parameters are primarily concentrated in the image encoder. Therefore, following previous work\cite{ma2024segment, MA-SAM, SAMed}, we employ low-rank adaptation (LoRA)\cite{hu2022lora} to fine-tune the image encoder. This allows the encoder to adapt to downstream tasks while keeping its parameters frozen. LoRA achieves parameter-efficient fine-tuning by performing low-rank decomposition on the parameters to be fine-tuned. For example, a pre-trained weight $W \in \mathbb{R}^{d \times k} $ can be updated as follows $\hat{W} = W + \triangle W = W + BA $, where $\hat{W} \in \mathbb{R}^{d \times k} $ denotes the updated weight matrix, and a low-rank decomposition $\triangle W = BA, B \in \mathbb{R}^{d \times r}, A \in \mathbb{R}^{r \times k}$. In our experiments, we use LoRA for every image encoder block and set the rank of LoRA, $r=32$ by default.

\subsection{Q\&A based Self-prompt Module}
To eliminate SAM's reliance on manual prompts, we designed the Q\&A prompt pairs to learn abstract task-guided prompts during the fine-tuning process. As shown in Figure 2, the Q\&A prompt pairs are learnable prompts that are concatenated with the input of the image encoder and the mask decoder, respectively. Q-prompts and A-prompts are paired, and their quantity is set to match the number of segmentation targets $c$. The Q\&A prompt pairs are connected through a bottleneck structure. It is worth noting that the bottleneck structure consists of a dimensionality reduction mapping function $f, f \in \mathbb{R}^{d_I \times d_D}$ and a task-specific MLP module. The task-specific MLP module, $\text{MLP} = \{mlp_1, mlp_2, ..., mlp_c\}$, contains $c$ independent $mlp$. Specifically, the relationship between the Q\&A prompt pairs can be formulated as follows. 
\begin{gather}
\begin{array}{ll}
    a_i = mlp_i(f(q_i)) & i = 1, 2, ..., c \\
\end{array}
\end{gather}

Namely, the Q-prompt, $Q = \{q_1, q_2, ..., q_c, q_i\in \mathbb{R}^{1 \times d_I}\}$ is randomly initialized to guide the image encoder to focus on key information relevant to segmentation,  while the A-prompt, $ A = \{a_1, a_2, ..., a_c, a_i \in \mathbb{R}^{1 \times d_D}\}$ guide the mask decoder to decode critical information from the corresponding encoded features. Where the $d_I,d_D$ represent the feature dimensions of the inputs to the image encoder and mask decoder, respectively. The Q\&A prompt pairs are established as question-answer-like relationship between the encoding and decoding processes. Optimizing the Q\&A prompt pairs effectively involves learning task-relevant cues.

\subsection{Hierarchical Feature Fusion Module}
Using a U-shaped structure to fuse high- and low-dimensional features effectively addresses common challenges in medical imaging. Therefore, we design a hierarchical feature fusion module to integrate features across different levels. The image encoder in vanilla SAM comprises local attention layers and global attention layers. 
As the encoding depth increases, the features captured by each global attention layer vary significantly. Therefore, distinct sets of Q\&A prompt pairs, mask decoder blocks, and bottlenecks are assigned for each layer rather than being shared across layers, building a hierarchical structure, which enables layer-wise decoding and feature fusion.
These modules are formally denoted as $\tilde{Q} = \{Q_1, Q_2, ..., Q_N, Q_i\in \mathbb{R}^{c \times d_I}\}$, $\tilde{A} = \{A_1, A_2, ..., A_N, A_i \in \mathbb{R}^{c \times d_D}\}$, $F = \{f_1, f_2, ..., f_N\}$, and $\text{MLPs} = \{\text{MLP}_1, \text{MLP}_2, ..., \text{MLP}_N\}$, where $N$ refers to the number of global attention layers in the image encoder. Hierarchical decoding blocks are also incorporated on the decoder side to facilitate progressive feature refinement. Additionally, skip connections are introduced to enhance the stability and convergence of the decoder across hierarchical levels. For each mask decoder module, the input features comprise the output from the corresponding global attention layer of the encoder, the output from the preceding mask decoder block, and the final output of the image encoder, which serves as the initial input for the entire decoding process. The proposed hierarchical feature fusion process can be formally expressed as follows.
\begin{gather}
\begin{array}{ll}
    A_j = \text{MLP}_j(f(Q_j))  & j = 1, 2, ..., N  \\
    output_N= \text{Decoder}_N(neck_N(embedding_N), A_N) &\\
    output_i= \text{Decoder}_i(output_{(i+1)}+neck_i(embedding_i)+output_N, A_i) &i = 1, 2, ..., N-1
\end{array}
\end{gather}
Here, $output_i$ denotes the output of the $i$-th mask decoder block, $\text{Decoder}_i$ represents the $i$-th mask decoder block, $neck_i$ refers to the neck projection that maps encoder image embeddings to decoder embeddings, and $embedding_i$ denotes the output of the $i$-th global attention layer in the image encoder. Note that the final layer of the image encoder is also a global attention layer; thus, the initial input to the mask decoder consists solely of $embedding_N$.

\subsection{Loss function}
Following previous medical image segmentation works \cite{MA-SAM,h-sam, Med-SA}, we design the overall loss of HSP-SAM as a weighted sum of Dice loss and cross-entropy loss, assigning weights to each:
\begin{gather}
L = \alpha L_{dice\_loss}(\hat{y}, y) + (1-\alpha)L_{ce\_loss}(\hat{y}, y) 
\end{gather}
where $\hat{y}$ represents the predicted output, $y$ denotes the ground truth, $L_{dice\_loss}$ and $L_{ce\_loss}$ correspond to the Dice loss and cross-entropy loss, respectively, and $\alpha$ is the weighting factor for the loss. In our experiments, we set $\alpha=0.8$.

%% file: secs/experiment.tex
\section{Experiments}
In this section, we aim to validate the performance of HSP-SAM on medical datasets without inputting any additional manual prompts. Section 4.1 and 4.2 assess its segmentation and generalization performance on polyp and skin lesion datasets. Section 4.3 examines its robustness across various medical modalities. Section 4.4 presents ablation studies on key modules, and Section 4.5 provides an in-depth model analysis.

\subsection{Polyp Segmentation}
\textbf{Setup.} Polyps are abnormal tissue growths in the gastrointestinal tract, some of which may become malignant. Accurate segmentation is crucial for early diagnosis but remains challenging due to their varied size, shape, texture, and contrast with surrounding tissues. To validate the effectiveness of HSP-SAM in this classic medical segmentation task, we follow the experimental setup of SAM-SP and evaluate on five polyp datasets including Kvasir-SEG \cite{kvasir}, CVC-ClinicDB\cite{clinicDB}, CVC-ClonDB\cite{colonDB}, Endoscene\cite{endoscene}, and ETIS-LaribDB\cite{etis}. In the experiments, we trained on a subset of CVC-ClinicDB and Kvasir datasets and evaluated the model across all datasets. Detailed experimental configurations are provided in Appendix A.1.

\textbf{Results.} Table 1 highlights the strong segmentation accuracy and generalization capability of HSP-SAM. For fair comparison, we report ViT-B backbone results (ViT-H is used elsewhere by default). Even with ViT-B, HSP-SAM outperforms existing prompt-free SAM-based methods, and with ViT-H, it surpasses many specialized models. On CVC-ClinicDB and Kvasir, HSP-SAM improves Dice scores by 2.78\% and 0.47\% over the best performer PolypPVT\cite{polyppvt}, respectively. In zero-shot tests on CVC-300, CVC-ColonDB, and ETIS-LaribDB, it achieves notable gains. compared to the best prompt-free SAM-based methods, HSP-SAM (ViT-B) achieves improvements of 1.28\%, 0.49\%, and 17.58\% on CVC-300, CVC-ColonDB, and ETIS-LaribDB, respectively. These results demonstrate HSP-SAM’s superior segmentation and generalization performance.

\begin{table}[tbph]
\vspace{-2mm}
    \centering
    \caption{Quantitative comparison on Polyp Segmentation of different approaches}
    \resizebox{\linewidth}{!}{
    \begin{tabular}{l|cc|cc|cc|cc|cc}
        \toprule
        \multirow{2}{*}{Models}  & \multicolumn{2}{c|}{CVC-ClinicDB} & \multicolumn{2}{c|}{Kvasir} & \multicolumn{2}{c|}{CVC-300} & \multicolumn{2}{c|}{CVC-ColonDB} & \multicolumn{2}{c}{ETIS-LaribDB} \\
        \cmidrule(lr){2-3} \cmidrule(lr){4-5} \cmidrule(lr){6-7} \cmidrule(lr){8-9}\cmidrule(lr){10-11} &DICE$\uparrow$ & IoU$\uparrow$ & DICE$\uparrow$ & IoU$\uparrow$ & DICE$\uparrow$ & IoU$\uparrow$ & DICE$\uparrow$ & IoU$\uparrow$ & DICE$\uparrow$ & IoU$\uparrow$ \\
        \midrule
        U-Net\cite{u-net} & 82.3 & 75.5 & 81.8 & 74.6 & 71.0 & 62.7 & 51.2 & 44.4 & 39.8 & 33.5 \\
        UNet++\cite{unet++}  & 79.4 & 72.9 & 82.1 & 74.3 & 70.7 & 62.4 & 48.3 & 41.0 & 40.1 & 34.4 \\
        PraNet\cite{pranet}  & 89.9 & 84.9 & 89.9 & 84.0 & 87.1 & 79.7 & 71.2 & 64.0 & 62.8 & 56.7 \\
        UACANet-L\cite{uacanet}  & 91.07 & 86.7 & 90.83 & 85.95 & 88.21 & 80.84 & 72.57 & 65.41 & 63.89 & 56.87 \\
        SSFormerPVT\cite{ssformerpvt}  & 92.88 & 88.27 & 91.11 & 86.01 & 89.46 & 82.68 & 79.34 & 70.63 & 78.03 & 70.1 \\
        PolypPVT\cite{polyppvt}  & 93.08 & \textbf{88.28} & 91.23 & 86.30 & 88.71 & 81.89 & 80.75 & 71.85 & 78.67 & 70.97 \\
        \midrule
        SAM\cite{sam}  & 33.29 & 25.64 & 61.48 & 53.74 & 45.00 & 38.62 & 29.33 & 31.23 & 24.76 & 20.37 \\
        SAMed\cite{SAMed}  & 83.34 & 76.32 & 88.01 & 81.61 & 83.63 & 76.27 & 70.57 & 62.69 & 60.13 & 52.05 \\
        SAM-Med2D\cite{sammed2d} & 85.91 & 80.60 & 87.06 & 80.88 & 84.81 & 77.49 & 69.08 & 60.79 & 59.80 & 53.00 \\
        Med-SA\cite{Med-SA} & 86.32 & 80.80 & 87.11 & 80.53 & 83.63 & 76.27 & 73.68 & 64.97 & 59.04 & 52.32 \\
        SAM-SP\cite{sam-sp} & 85.91 & 80.37 & 90.57 & 85.46 & 88.94 & 82.55 &74.67 & 67.24 & 64.87 & 58.11 \\
        H-SAM\cite{h-sam} &70.21 & 55.61 & 79.45& 68.54 & 79.6 & 68.27 & 62.23 & 50.63 & 56.58 & 32.98 \\
        \midrule
        \textbf{HSP-SAM(ViT-B)} & 89.91 & 82.02 & 89.62 & 84.06  & 90.22 & 82.43& 75.16 & 67.76 & 82.45 & 67.16\\
        \textbf{HSP-SAM(ViT-H)} & \textbf{95.86} & 87.98  & \textbf{91.70} & \textbf{86.37} & \textbf{90.83} & \textbf{84.73} & \textbf{81.61} & \textbf{73.78} & \textbf{84.91} & \textbf{71.63} \\
        \bottomrule
    \end{tabular}
    }
\end{table}

\subsection{Skin Lesion Segmentation}

\begin{table}
\vspace{-15mm} 
\renewcommand{\arraystretch}{0.8}
    \centering
     \caption{\small Quantitative comparison on Skin Lesion Segmentation of different approaches}
    \resizebox{0.5\linewidth}{!}{
    \begin{tabular}{lcccc}
        \toprule
        \multirow{2}{*}{Methods} & \multicolumn{2}{c}{ISIC2017} & \multicolumn{2}{c}{ISIC2018} \\
        \cmidrule(lr){2-3} \cmidrule(lr){4-5}
        & DICE$\uparrow$ & IoU$\uparrow$ & DICE$\uparrow$ & IoU$\uparrow$ \\
        \midrule
        U-Net\cite{u-net} & 86.99 & 76.98 & 87.55 & 77.86 \\
        UNet++\cite{unet++}  & 82.10 & 72.41 & 87.83 & 78.31 \\
        TransFuse\cite{zhang2021transfuse}  & 88.40 & 79.21 & 89.27 & 80.63 \\
        MALUNet\cite{ruan2022malunet} & 88.13 & 78.78 & 89.04 & 80.25 \\
        EGE-Unet\cite{ege-unet}  & 88.77 & 79.81 & 89.04 & 80.25 \\
        \midrule
        SAM\cite{sam} & 53.28 & 40.80 & 58.79 & 46.06 \\
        SAM-Med2D\cite{sammed2d} & 87.01 & 79.63 & 88.92 & 81.87\\
        SAMed\cite{SAMed} & 85.86 & 77.51 & 88.73 & 79.94 \\
        Med-SA\cite{Med-SA} & 87.78 & \textbf{80.35} & 88.75 & 81.77 \\
        SAM-SP\cite{sam-sp}  & 87.66 & 80.01 & 89.39 & \textbf{82.31} \\
        H-SAM\cite{h-sam} & 83.80 & 69.72& 87.04 & 76.80 \\
        \midrule
        \textbf{HSP-SAM(ViT-B)} & \textbf{89.21} & 78.81 & \textbf{90.32} & 81.78 \\
        \bottomrule
    \end{tabular}
    
    }
    \vspace{-8mm} 
\end{table}
\textbf{Setup.}
Skin lesion segmentation is another classic and well-established medical image segmentation task. Compared to other domains, it benefits from abundant annotated data and a series of mature public competitions, such as the ISIC challenges. In this experiment, we further validate the performance of HSP-SAM on the classic skin cancer segmentation task using the ISIC 2017\cite{2017isic} and ISIC 2018\cite{isic2018} datasets. To ensure a fair comparison, we follow the 7:3 training-to-test split strategy adopted in previous studies\cite{sam-sp}. Details of the datasets are provided in Appendix A.2.

\textbf{Results.} Table 2 reports the segmentation performance of HSP-SAM on skin lesion datasets. Following the same protocol as in polyp segmentation, we use the ViT-B backbone for fair comparison. HSP-SAM achieves improvements of 0.44\% and 0.93\% in Dice score over the best existing methods on ISIC-2017 and ISIC-2018, respectively. Although the margins are relatively small, these results further demonstrate the effectiveness and versatility of HSP-SAM as a unified framework for multi-task medical image segmentation.
\vspace{-2mm}

\subsection{Multi Modalities Generalization Evaluation}
\textbf{Setup.}
Medical image segmentation is inherently more complex than natural image segmentation, largely due to the diversity of imaging modalities and segmentation tasks. Traditional automated medical segmentation models are often specialized architectures designed to address this complexity. To validate that HSP-SAM serves as a modality-agnostic and highly generalizable unified framework, we collected five types of medical imaging modalities, each comprising a source and a target dataset. HSP-SAM is trained on the source datasets and evaluated on both the source and target datasets. Evaluations on the target datasets correspond to a zero-shot setting. For dataset partitioning, we maintained consistency with the prior studies\cite{esp-sam}. For datasets lacking established splits, we adopted a 7:3 ratio for the training-test division. Detailed data specifications are provided in Appendix A.3.

\textbf{Results.} 
Tables 4 and 5 comprehensively evaluate the segmentation and generalization capabilities of HSP-SAM across a wide range of medical imaging modalities. Table 4 presents the results on source datasets after training, where HSP-SAM consistently achieves performance comparable to or exceeding that of state-of-the-art (SOTA) methods, with the only minor exception being the DRIVE vessel segmentation task. Notably, the advantages of HSP-SAM become even more pronounced in Table 5, which reports zero-shot transfer performance on unseen target datasets. Specifically, HSP-SAM achieves remarkable improvements of 14.04\%, 7.91\%, and 10.36\% in Dice scores on the $\mathcal{T}{3}$, $\mathcal{T}{4}$, and $\mathcal{T}{5}$ datasets, respectively, significantly outperforming the best existing models. These results demonstrate not only the strong segmentation ability of HSP-SAM but also its exceptional generalization across different medical modalities and imaging distributions. Furthermore, HSP-SAM consistently achieves the best results on the Hausdorff Distance (HD) metric across all datasets, highlighting its ability to produce more precise boundary delineations.

\begin{table}[h!]
\centering
\captionsetup{justification=centering}
\caption{Comparison with State-of-the-Art Frameworks in Universal Medical Image Segmentation (source domains).} 
 \resizebox{1.0\textwidth}{!}{
    \begin{tabular}{l c c c c c c c c c c c}
    \toprule
     \multirow{2}{*}{Methods} & \multirow{2}{*}{Manual Prompt} & \multicolumn{2}{c}{ $\mathcal{S}^{1}$} & \multicolumn{2}{c}{ $\mathcal{S}^{2}$} & \multicolumn{2}{c}{ $\mathcal{S}^{3}$} & \multicolumn{2}{c}{ $\mathcal{S}^{4}$} & \multicolumn{2}{c}{ $\mathcal{S}^{5}$} \\
     \cmidrule(lr){3-12}
     & & DICE$\uparrow$ & HD$\downarrow$ & DICE$\uparrow$ & HD$\downarrow$  & DICE$\uparrow$ & HD$\downarrow$ & DICE$\uparrow$ & HD$\downarrow$ & DICE$\uparrow$ & HD$\downarrow$ \\
    \midrule
    U-Net\cite{u-net} & & 82.87 & 180.90  & 79.13 & 68.22 & 83.91 & 130.96 & 69.24 & 131.62 & 88.16 & 130.28 \\
     U-Net++\cite{unet++}  & & 82.69 & 175.04  & 80.61 & 65.03 & 85.77 & 152.96 & 72.50 & 137.84 & 90.48 & 112.12 \\
     Att-UNet\cite{att-unet}  & & 83.97 & 170.78 & 80.72 & 66.48 & 86.90 & 156.72 & 71.02 & 107.70 & 91.12 & 113.64 \\
     nnUNet\cite{isensee2021NnUNet}  & & 84.96 & 126.19 & 81.71 & 64.16 & 88.38 & 127.28 & 75.22 & 119.66 & 91.61 & 121.16 \\
     H2Former\cite{he2023h2former}  &  \ding{55} & 82.12 & 191.39 & 81.46 & 65.09 & 84.66 & 142.08 & 70.30 & 110.98 & 90.17 & 117.69 \\
     TransUNet\cite{chen2021transunet} & & 84.28 & 134.80 & 81.68 & 64.91 & 86.00 & 151.14 & 71.07 & 123.80 & 90.03 & 109.56 \\
     ADS\cite{adc}  & & 84.14 & 172.84 & 80.48 & 68.49 & 87.70 & 117.74 & 72.55 & 136.08 & 90.32 & 115.24 \\
     CIAug\cite{ouyang2022causality}  & & 83.91 & 141.07 & 80.45 & 65.72 & 87.69 & 106.68 & 71.78 & 134.84 & 90.58 & 113.72 \\
     MADGNet\cite{nam2024modality}  & & 85.02 & 131.84 & 81.89 & 64.73 & 88.20 & 107.16 & 72.75 & 131.24 & 91.38 & 98.04 \\
     \midrule 
    MobileSAM\cite{mobilesam} & & 87.97 & 105.34 & 69.31 & 94.41 & 81.83 & 86.46 & 66.48 & 107.78 & 87.42 & 131.00 \\
     RepViT-SAM\cite{wang2023repvit} & & 88.00 & 106.75  & 67.76 & 98.66 & 81.81 & 154.65 & 68.38 & 103.09 & 88.81 & 127.17 \\
     EfficientViT-SAM\cite{zhang2024efficientvit} &  \multirow{2}{*}{Point} & 88.49 & 103.61 & 78.16 & 77.16 & 85.16 & 102.72 & 74.71 & 113.18 & 89.37 & 116.42 \\
     EfficientSAM\cite{xiong2024efficientsam} &  & 87.11 & 108.12  & 76.32 & 79.41 & 82.81 & 96.35 & 71.17 & 113.57 & 88.41 & 129.08 \\
     EdgeSAM\cite{zhou2023edgesam} & & 88.10 & 100.92  & 68.04 & 92.40 & 81.76 & 105.06 & 67.64 & 105.51 & 87.35 & 113.46 \\
     SAM-Lightening\cite{song2024samligtning}  & & 88.28 & 101.64 & 74.84 & 87.32 & 83.70 & 97.79 & 73.07 & 129.53 & 89.18 & 111.67 \\
    \midrule
     MobileSAM\cite{mobilesam} & & 86.19 & 168.48  & 25.56 & 254.24 & 77.29 & 284.32 & 61.51 & 342.22 & 61.49 & 307.44 \\
     RepViT-SAM\cite{wang2023repvit}  & & 85.73 & 157.66 & 25.10 & 263.76 & 78.75 & 296.36 & 61.62 & 440.75 & 60.54 & 287.63 \\
     EfficientViT-SAM\cite{zhang2024efficientvit}  &  & 87.18 & 151.29 & 25.95 & 252.39 & 82.48 & 317.51 & 65.44 & 354.98 & 66.32 & 292.69 \\
     EfficientSAM\cite{xiong2024efficientsam}  & \multirow{2}{*}{\ding{55}} & 87.02 & 162.45 & 25.84 & 261.51 & 78.54 & 305.36 & 62.05 & 430.18 & 60.56 & 336.55 \\
     EdgeSAM\cite{zhou2023edgesam} & & 85.86 & 153.97 & 25.62 & 257.85 & 76.41 & 328.79 & 60.24 & 400.69 & 59.68 & 357.20 \\
     SAM-Lightening\cite{song2024samligtning} & & 86.99 & 165.80 & 25.52 & 261.25 & 80.14 & 292.33 & 62.59 & 376.55 & 62.68 & 306.66 \\
     H-SAM\cite{h-sam} & & 88.88 & \textbf{20.43} & - & - & 81.73 & 43.23 & 79.66 & 31.00 & 86.52 & \textbf{32.28} \\ 
    ESP-MedSA\cite{esp-sam} &  & 88.52 & 92.42  & \textbf{82.42} & 62.64 & 92.93 & 56.32 & 85.28 & 82.32 & \textbf{92.24} & 85.93 \\
    \midrule
    HSP-SAM & \ding{55} & \textbf{89.17} & 38.08 & 78.9 & \textbf{32.44} & \textbf{94.82} & \textbf{23.87} & \textbf{89.19} & \textbf{20.20} & 92.19 & 38.63 \\
    \bottomrule
    \end{tabular}
    }
\end{table}

\begin{table}[h!]
    \centering
    \captionsetup{justification=centering}
    \caption{Comparison with State-of-the-art Frameworks in Domain-Generalized Medical Image Segmentation (unseen domains).}
    \resizebox{1.0\textwidth}{!}{
    \begin{tabular}{l c c c c c c c c c c c}
    \toprule
    \multirow{2}{*}{Methods} & \multirow{2}{*}{Manual Prompt} & \multicolumn{2}{l}{ $\mathcal{S}^{1}\rightarrow \mathcal{T}^{1}$ } & \multicolumn{2}{l}{ $\mathcal{S}^{2} \rightarrow \mathcal{T}^{2}$} & \multicolumn{2}{l}{ $\mathcal{S}^{3} \rightarrow \mathcal{T}^{3}$ } & \multicolumn{2}{l}{ $\mathcal{S}^{4} \rightarrow \mathcal{T}^{4}$ } & \multicolumn{2}{l}{ $\mathcal{S}^{5} \rightarrow \mathcal{T}^{5}$ }  \\
    \cmidrule(lr){3-12}
    & & Dice  $\uparrow$  & HD  $\downarrow$  & Dice  $\uparrow$  & HD  $\downarrow$  & Dice  $\uparrow$  & HD  $\downarrow$  & Dice  $\uparrow$  & HD  $\downarrow$  & Dice  $\uparrow$  & HD  $\downarrow$ \\
    \midrule 
    U-Net\cite{u-net} & & 87.00 & 140.38 & 61.21 & 174.97 & 32.60 & 461.64 & 39.17 & 336.98 & 46.57 & 295.62 \\
    U-Net++\cite{unet++} & & 87.87 & 110.08& 62.79 & 180.01 & 36.43 & 466.92 & 41.30 & 343.66 & 47.77 & 281.42 \\
     Att-UNet\cite{att-unet} & & 88.66 & 126.35 & 65.18 & 118.10 & 35.56 & 420.60 & 41.89 & 357.08 & 48.70 & 280.56 \\
     nnUNet\cite{isensee2021NnUNet} & & 89.56 & 105.53 & 65.00 & 107.82 & 36.00 & 490.92 & 43.87 & 269.04 & 49.19 & 283.67 \\
     H2Former\cite{he2023h2former} &  \ding{55}  & 87.79 & 146.24 & 65.68 & 115.71 & 34.72 & 480.78 & 42.46 & 275.80 & 53.86 & 294.98 \\
     TransUNet\cite{chen2021transunet} & & 89.26 & 108.48 & 66.23 & 112.06 & 42.79 & 339.14 & 44.84 & 267.26 & 54.22 & 282.64 \\
     ADS\cite{adc} & & 87.83 & 129.46& 62.45 & 172.16 & 37.36 & 454.70 & 43.19 & 276.30 & 51.06 & 281.34 \\
     CIAug\cite{ouyang2022causality} & & 88.44 & 127.62 & 65.65 & 120.87 & 39.07 & 387.02 & 41.50 & 267.40 & 53.92 & 286.23 \\
     MADGNet\cite{nam2024modality} & & 89.71 & 96.86& 66.88 & 119.41 & 44.32 & 365.84 & 44.91 & 264.61 & 59.29 & 278.56 \\
     \midrule
     MobileSAM\cite{mobilesam} & & 90.37 & 87.28 & 54.10 & 160.16 & 32.77 & 399.29 & 38.68 & 310.52 & 16.17 & 452.18 \\
     RepViT-SAM\cite{wang2023repvit} & & 90.63 & 84.74  & 55.72 & 133.82 & 27.76 & 381.39 & 33.58 & 301.85 & 15.20 & 438.64 \\
     EfficientViT-SAM\cite{zhang2024efficientvit} &\multirow{2}{*}{Point} & 91.14 & 85.05  & 72.12 & 116.89 & 61.67 & 179.99 & 58.58 & 183.63 & 34.24 & 332.94 \\
     EfficientSAM\cite{xiong2024efficientsam} &  & 90.80 & 89.34  & 69.20 & 98.05 & 56.50 & 218.22 & 52.95 & 233.98 & 25.14 & 358.13 \\
     EdgeSAM\cite{zhou2023edgesam} & & 90.38 & 86.32 & 56.03 & 136.48 & 28.17 & 433.08 & 37.51 & 291.65 & 12.11 & 501.14 \\
     SAM-Lightening\cite{song2024samligtning} & & 90.85 & 89.96  & 67.38 & 98.85 & 58.12 & 210.20 & 54.75 & 244.37 & 23.69 & 423.15 \\
     \midrule
     MobileSAM\cite{mobilesam} & & 85.61 & 304.75  & 2.23 & 253.99 & 20.90 & 470.94 & 31.03 & 342.10 & 6.44 & 375.52 \\
     RepViT-SAM\cite{wang2023repvit} & & 84.69 & 283.65  & 2.30 & 283.22 & 17.64 & 491.23 & 26.64 & 409.92 & 5.96 & 383.02 \\
     EfficientViT-SAM\cite{zhang2024efficientvit} &  & 89.29 & 177.12  & 9.72 & 256.58 & 58.77 & 357.63 & 46.08 & 501.00 & 14.85 & 442.26 \\
     EfficientSAM\cite{xiong2024efficientsam} &  \ding{55}  & 88.44 & 281.54 & 6.75 & 298.82 & 46.99 & 474.12 & 39.44 & 468.81 & 12.81 & 419.71 \\
     EdgeSAM\cite{zhou2023edgesam} & & 84.31 & 334.93 & 2.37 & 266.70 & 17.64 & 489.03 & 30.01 & 382.82 & 5.83 & 476.44 \\
     SAM-Lightening\cite{song2024samligtning} & & 89.41 & 168.96  & 4.26 & 251.03 & 56.94 & 419.96 & 43.68 & 446.36 & 12.01 & 439.31 \\
     H-SAM\cite{h-sam} & & 91.34 & 19.43 & - & - & 62.52 & 69.84 & 54.63 & 70.94 & 35.94 & 61.03 \\
     ESP-MedSAM\cite{esp-sam} &    & 91.45 & 78.13 & \textbf{79.68} & 83.84 & 65.96 & 160.13 & 61.62 & 199.74 & 64.21 & 248.47 \\
    \midrule
    HSP-SAM & \ding{55} & \textbf{92.05} & \textbf{36.23} & 77.73 & \textbf{42.69} & \textbf{80.00} & \textbf{67.34} & \textbf{69.53} & \textbf{82.24} & \textbf{74.57} & \textbf{86.72}\\
    \bottomrule
    \end{tabular}
}
    \label{tab:my_label}
\end{table}






\subsection{Ablation Experiments}
\textbf{Setup.} To validate the effectiveness and generalization capability of our proposed modules--Q\&A-based self-prompting module, hierarchical decoding module, and skip connections module, we conducted ablation studies on Task 3, with fine-tuned SAM serving as the baseline model. We conducted 5 ablation experiments with the results summarized in Table 6. The first variant (Ablation\_1) implemented Q\&A prompt pairs. Then we introduce the hierarchical decoding structure for Ablation\_2. In Ablation\_3 we further enhanced this architecture by integrating skip connections. For comparative analysis, Ablatio\_4 maintained the hierarchical decoding structure while removing both Q\&A prompt pairs and skip connections, whereas Ablation\_5 preserved the hierarchy and skip connections but eliminated the Q\&A prompt pairs. This structured ablation study enables precise attribution of performance gains to individual architectural innovations.

\textbf{Results.} 
As shown by the ablation results, the progressive addition of the three modules leads to consistent performance improvements. However, interactions between the modules exist, and their combined effect is not simply additive compared to the baseline. The ablation study indicates that integrating all three modules yields the best overall performance.
Therefore, all experimental results reported in this paper are based on the Ablation\_3 configuration.

\begin{table}[h!]
    \centering
    \caption{Ablation study of HSP-SAM in domain-generalized Medical Image Segmentation:  $\mathcal{S} \rightarrow \boldsymbol{\mathcal { T }}$}
   \resizebox{1.0\textwidth}{!}{
    \begin{tabular}{cccc|ccc}
        \toprule
        Module & Q\&A pairs & Hierarchical decoding & Skip connection & Dice (Avg.)  $\uparrow$  & HD (Avg.)  $\downarrow$ & Params(M) $\downarrow$\\
        \midrule
        Ft-SAM & & & & 64.49 & 451.43 & 90.58 \\
        Ablation\_1 & \checkmark & & &74.28 & 70.89 & 3.4\\
        Ablation\_2 & \checkmark &  \checkmark  & & 76.81 & 63.55 & 23.96 \\
        Ablation\_3 & \checkmark & \checkmark &  \checkmark  & \textbf{78.78}  & 63.04  & 23.96\\
        Ablation\_4 &  &  \checkmark  & & 77.87 & \textbf{61.87}  & 21.27\\
        Ablation\_5 &  & \checkmark &  \checkmark  & 77.81 & 63.66  & 21.27\\
        \bottomrule
    \end{tabular}
}
    \label{tab:my_label}
\end{table}

\subsection{Model Analysis}

\textbf{Prompt Quantity Analysis.} In the original SAM framework, segmentation performance heavily relies on manually provided prompts that explicitly indicate the segmentation targets, with the number of prompts typically corresponding to the number of target objects. This study aims to investigate whether the abstract task-guided prompts generated through our proposed self-prompting approach are closely associated with specific segmentation targets or capture a broader understanding of the segmentation task itself, and how the number of prompts influences the model’s segmentation performance. Specifically, in this experiment, we follow the same multi-modality generalization settings. We evaluate the model’s performance under consistent training conditions with prompt quantities of 1, 2, 4, 8, and 16.

\begin{figure}[h!]
    \centering
    \includegraphics[width=1.0\linewidth]{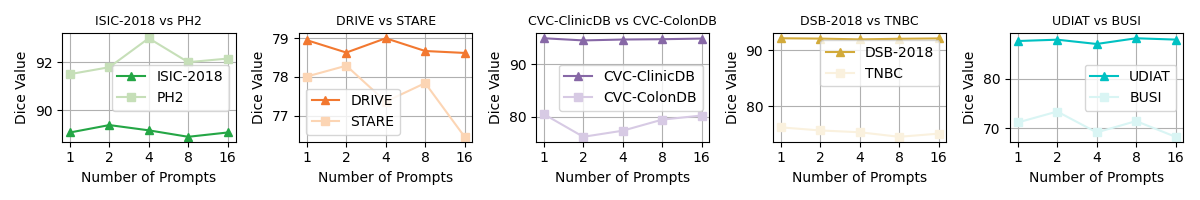}
    \caption{\textbf{Prompt Quantity vs. Segmentation Accuracy Across Datasets.} Results on the source datasets are marked with triangles, and results on the target datasets are marked with squares.}
    \label{fig:enter-label}
\end{figure}
Fig. 3 shows that segmentation performance on source datasets remains stable despite changes in the number of abstract task-guided prompts. This demonstrates that our hierarchical Q\&A self-prompting approach enables the model to learn high-level task representations rather than positional cues, achieving better generalization without requiring one-to-one correspondence with segmentation targets.
The DSB-2018 dataset serves as compelling evidence supporting this observation. Given that each input in this nuclei segmentation task contains a large number of targets (typically ranging from tens to dozens), if the abstract task-guided prompts were closely associated with individual objects, one would expect a clear performance improvement with an increased number of prompts. However, the segmentation performance remains largely invariant across different prompt quantities, indicating that the learned prompts encapsulate a higher-level understanding of the segmentation task rather than relying on explicit correspondence to each target.
Additionally, segmentation performance on target datasets shows slightly larger but non-monotonic fluctuations with varying numbers of abstract task-guided prompts. We attribute these fluctuations to normal variations in model performance under the zero-shot setting on unseen datasets.

\begin{figure}[h]
    \centering
    \includegraphics[width=1\linewidth]{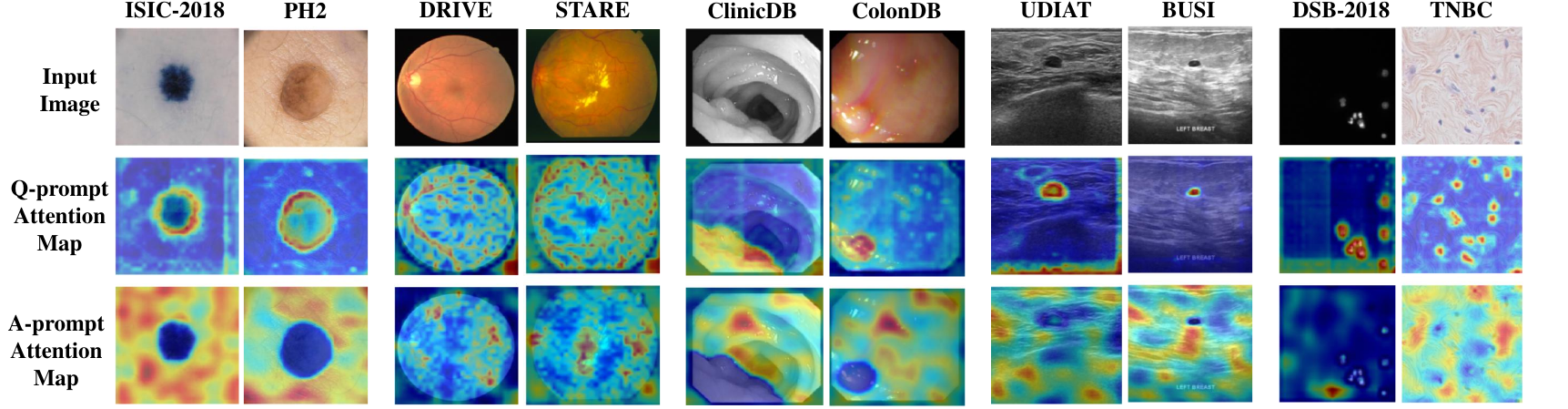}
    \caption{Visualization of attention heatmaps of Q\&A prompt pairs}
    \label{fig:enter-label}
\end{figure}
\textbf{Q\&A Prompt Pairs Analysis.}
Fig. 4 shows the attention heatmaps of Q and A prompt pairs with respect to the input images. The abstract task-guided prompts are implemented as Q\&A prompt pairs within the model, and Fig. 4 intuitively illustrates their role in guiding the segmentation process. We observe that the Q prompts primarily attend to the foreground objects, while the A prompts focus more on background regions. This behavior likely reflects their respective roles: Q prompts guide the image encoder to emphasize task-relevant areas, whereas the A prompts facilitate the decoding process, requiring attention to both the background and target structures. Notably, without explicit positional inputs, our proposed Q\&A prompt pairs still effectively capture the spatial distribution of segmentation targets and even outperform conventional positional prompts on some datasets. While positional prompts are insufficient for vessel segmentation (DRIVE and STARE) and become impractical for nucleus segmentation (DSB-2018 and TNBC) due to the large number of targets, our abstract task-guided prompts consistently achieve accurate target localization without manual inputs across diverse datasets. The comparison results on source and target datasets demonstrate that the proposed abstract task-guided prompts capture a deep understanding of the segmentation task itself. This enables the model to maintain strong performance across target datasets that, despite sharing the same modality, differ significantly in the number of segmentation targets and overall image quality.


%% file: secs/conclusion.tex
\section{Conclusion}

We propose HSP-SAM, a prompt-free medical image segmentation framework via hierarchical self-prompting method. 
By leveraging Q\&A prompt pairs, it autonomously learns abstract task-guided prompts during fine-tuning, enabling strong segmentation and generalization performance across diverse medical imaging modalities. To the best of our knowledge, we are the first to abandon positional prompts in self-prompting approaches aimed at mitigating SAM's dependence on manual inputs. We observe that the abstract task-guided prompts generated through self-prompting offer a higher-level understanding of the overall segmentation task, beyond the specific, input-dependent guidance provided by conventional positional prompts. This broader task awareness enables significantly stronger generalization across medical datasets. We believe that such abstract task understanding, which can bridge concrete visual features and abstract linguistic representations, may pave the way toward universal medical image segmentation models.